\documentclass{ifacconf}

\usepackage{natbib}        
\usepackage{adjustbox}
\usepackage{amsmath,amsfonts, mathtools, amssymb, multirow}
\usepackage{array}

\usepackage{textcomp}
\usepackage{stfloats}
\usepackage{url}
\usepackage{verbatim}
\usepackage{graphicx}

\usepackage{subcaption, multirow}

\makeatletter
\let\old@ssect\@ssect 
\makeatother

\usepackage{hyperref}

\makeatletter
\def\@ssect#1#2#3#4#5#6{%
  \NR@gettitle{#6}%
  \old@ssect{#1}{#2}{#3}{#4}{#5}{#6}%
}
\makeatother

\hypersetup{
	colorlinks=true,
	linkcolor=blue,
	citecolor=blue,
	urlcolor=blue,
}

\makeatletter
\def\blfootnote#1{\add@tok\t@glob@notes{\begingroup\def\@thefnmark{}\footnotetext{#1}\endgroup}}
\makeatother

\begin{document}

\begin{frontmatter}

\title{Mobile Robot Exploration Without Maps via Out-of-Distribution Deep Reinforcement Learning} 

\blfootnote{Digital Object Identifier (DOI): 10.1016/j.ifacol.2025.12.292}
\blfootnote{© 2025 the authors. This work has been accepted to IFAC for publication under a Creative Commons License CC-BY-NC-ND.}

\author[First]{Shathushan Sivashangaran} 
\author[Second]{Apoorva Khairnar} 
\author[First]{Azim Eskandarian}

\address[First]{Virginia Commonwealth University, 
   Richmond, VA 23284 USA (e-mail: sivashangars@vcu.edu; eskandariana@vcu.edu)}
\address[Second]{Virginia Tech, Blacksburg, VA 24061, USA (e-mail: apoorvak@vt.edu)}

\begin{abstract}                
Autonomous Mobile Robot (AMR) navigation in dynamic environments that may be GPS denied, without a-priori maps, is an unsolved problem with potential to improve humanity’s capabilities. Conventional modular methods are computationally inefficient, and require explicit feature extraction and engineering that inhibit generalization and deployment at scale. We present an Out-of-Distribution (OOD) Deep Reinforcement Learning (DRL) approach that includes functionality in unstructured terrain and dynamic obstacle avoidance capabilities. We leverage accelerated simulation training in a racetrack with a transition probability to parameterize spatial reasoning with intrinsic exploratory behavior, in a compact, computationally efficient Artificial Neural Network (ANN), which we transfer zero-shot with a reward component to mitigate differences between simulation and real world physics. Our approach enables utility without a separate high-level planner or real-time cartography and utilizes a fraction of the computation resources of modular methods, enabling execution in a range of AMRs with different embedded computer payloads.
\end{abstract}


\end{frontmatter}

\section{INTRODUCTION} \label{se:introduction}

Autonomous Mobile Robots (AMRs) such as wheeled vehicles, quadrupeds, and humanoids are beneficial for a range of tasks in agriculture, manufacturing, disaster response, Search and Rescue (SAR), military, and extraterrestrial exploration. Operation in new, dynamic, GPS-denied environments without prior maps such as caves, lava tubes on Mars, unknown buildings, contested regions, or disaster areas remains an unsolved problem (\cite{zghair2021one}).

Current state-of-the-art approaches use modular methods that incorporate Simultaneous Localization and Mapping (SLAM) (\cite{cadena2016SLAMsurvey}) to estimate state via onboard sensors and build real-time environment models for collision-free trajectory planning and control. While integrating Machine Learning (ML) methods like Deep Reinforcement Learning (DRL) with SLAM has enabled effective exploration of unknown environments with static obstacles (\cite{hu2021sim, gervet2023navigating}), SLAM's reliance on sequential motion estimation prone to mapping and pose errors (\cite{murai2024mast3r}) is a major drawback, especially during initial exploration of new environments, which is the most challenging. Furthermore, conventional motion planning algorithms depend on specific configurations, limiting adaptability in information-poor, dynamic environments. 

Modular methods are computationally expensive, requiring multiple Artificial Neural Networks (ANNs) per subtask, making these potentially infeasible for real-time high-speed applications or resource-constrained small AMRs like the Multi-Modal Mobility Morphobot (\cite{sihite2023multi}) tested by NASA for Mars exploration. Comparatively, end-to-end navigation mapping sensor inputs directly to controls is efficient and eliminates explicit feature extraction (\cite{chib2023recent}). Most end-to-end AMR navigation works such as autonomous driving use vision-based Deep Learning (DL) with Convolutional Neural Networks (CNNs) and Vision Transformers (ViTs) due to DRL challenges like sample inefficiency and reward formulation, however DL requires millions of miles of curated, labeled data, which is labor-intensive.

Unlike DL, DRL requires no curated data, can utilize simple two-layer networks for parameter efficient navigation training via actor-critic algorithms (\cite{song2023reaching, sivashangaran2023deep}), and can learn superhuman policies beyond human knowledge, as seen in Artificial Intelligence (AI) surpassing humans in games and drone racing (\cite{kaufmann2023champion, schrittwieser2020mastering}). Hence, a ubiquitous end-to-end DRL model adaptable to new environments without reprogramming or maps, functional across AMR forms from small Autonomous Ground Vehicles (AGVs) to humanoids with varying embedded compute payload capacities, has potential to enhance AMR utility and deployment in dynamic real world settings.

We propose a novel DRL approach for end-to-end AMR navigation, trained with fewer samples in simulation using a racetrack environment transition probability for Out-of-Distribution (OOD) generalization to varied new environments without prior maps or state estimation for real-time cartography, transferring zero-shot to the real world via a reward component mitigating sim-to-real physics differences, relying solely on onboard sensors for GPS-denied functionality without a high-level planner for explicit exploration goals, and using far less computation than modular methods. Our contributions can be summarized as follows:

\begin{itemize}

\item We formulate a DRL framework and develop simulation training assets for sim-to-real end-to-end AMR exploration without a-priori maps.

\item We evaluate different combinations of DRL reward formulations, algorithms, and environment transition probabilities to shed light on the significance of formulating strategic training conditions for efficient learning, OOD capabilities and real world transfer. 

\item Spatial reasoning, with intrinsic exploratory behavior, dynamic obstacle avoidance capabilities, and robustness in unstructured terrain, is parameterized in a single, computationally efficient ANN. We show that DRL sample efficiency increases significantly with tactical training in a constrained racetrack environment at physical limits for OOD generalization as opposed to training in object-filled environments similar to the intended application. 

\end{itemize}

\section{RELATED WORK}

Most prior works use SLAM to estimate environmental landmarks for informed path planning (\cite{wen2020path}), while recent methods combine ML with SLAM using multiple ANNs for mapping, planning, and control. \cite{gervet2023navigating}) achieved navigation in unknown static indoor environments by predicting maps and agent pose via neural SLAM, computing long-term goals with a DRL trained CNN global policy to maximize coverage, and deriving actions via a RNN local policy from RGB observations and short-term goals, outperforming an end-to-end approach that used high-dimensional image input by 67\% due to the large image domain gap between the simulator used for training, and the real world. An alternative method that did not utilize SLAM leveraged a pre-trained ViT with a diffusion policy for undirected exploration and goal-conditioned actions, evaluated on smooth surfaces at low speeds (\cite{sridhar2023nomad}), but both state-of-the-art methods were computationally intensive, hindering high-speed operation, and neither demonstrated unstructured terrain traversal or dynamic obstacle avoidance.

Previous unstructured terrain navigation methods extracted traversability features via geometric and statistical processing (\cite{papadakis2013terrain}), with learning-based methods classifying terrain for traversability (\cite{schilling2017geometric}) to enable real world application without explicit rules, though requiring large manually labeled data and yielding low accuracy from discretization simplifications. More recent works added traversability modules to SLAM, such as in (\cite{hu2021sim}), where a DRL model trained in simulation and deployed in real world lab rough terrain, computed high-level discretized actions, that included forward or backward motion for specific distances and left or right rotation angles, using AMR pose and a 2D elevation map derived from a 3D mesh reconstructed via Poisson surface reconstruction from SLAM generated 3D point clouds.

\section{METHODOLOGY} \label{se:method}

We used Proximal Policy Optimization (PPO) (\cite{schulman2017proximal}), an on-policy model-free actor-critic Reinforcement Learning (RL) algorithm to train the policy $\pi(a\mid \mathcal{O})$, to compute continuous actions in action space $a$ for continuous observations in observation space $\mathcal{O}$, to maximize expected future rewards utilizing an ANN parameterized with weights $\theta$. The actor and critic networks each comprise 2 fully connected layers with 64 nodes each, with $tanh$ activation, that share parameters for efficient computation. 

\begin{figure}[!h]
    \centering
    \includegraphics[width = 0.4\columnwidth]{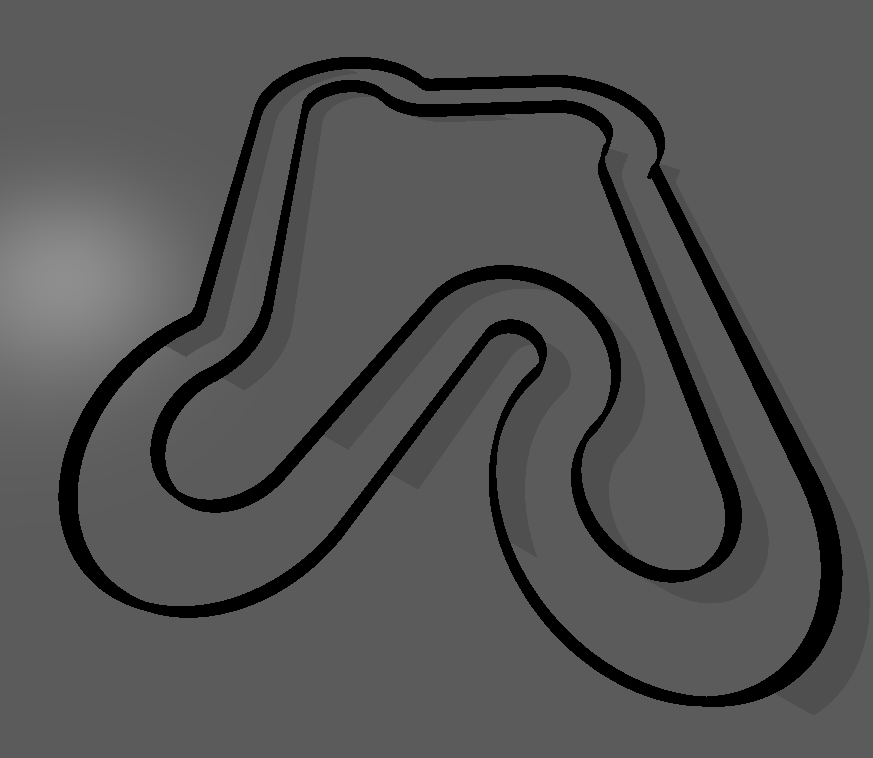}
    \caption{Multidirectional racetrack used for OOD training.} \label{racetrack_sim}
\end{figure}

The Markov Decision Process (MDP) mathematical framework used in RL comprises observations $\mathcal{O}$, actions $a$, transition probability $P(\mathcal{O}'\mid\mathcal{O}, a)$ which defines the likelihood of the next observation $\mathcal{O}'$ for action $a$ in observation $\mathcal{O}$, reward function $R$, and discount factor $\gamma$. The transition probability represents the environment dynamics that is unknown to the agent in model-free RL. In order to efficiently parameterize spatial reasoning, we trained the DRL agent to time-optimally navigate a multi-directional racetrack with varying cornering radii, as illustrated in Figure \ref{racetrack_sim}, which requires action computation at physical limits. We found the transition probability in this training environment with these conditions, to improve sample efficiency, obstacle avoidance success rate and intrinsic exploratory behavior over more intuitive obstacle cluttered environments similar to the intended AMR exploration applications, that we investigated in prior work (\cite{sivashangaran2023deep}). 

The training environment comprises a mixture of turning directions and lengths that prevents the model from overfitting to a specific layout, enabling OOD generalization, by providing an array of observation sequences for learning of diverse obstacle avoidance behavior at the maximum attainable velocity within the laws of physics. This requires the learning of a policy that plans action sequences dependent on the profiles of detected objects, which facilitates dynamic obstacle avoidance capabilities, and promotes the parameterization of intrinsic exploratory behavior with sparse rewards. 

\begin{figure*}[!h]
    \centering
    \includegraphics[width = 0.7\textwidth]{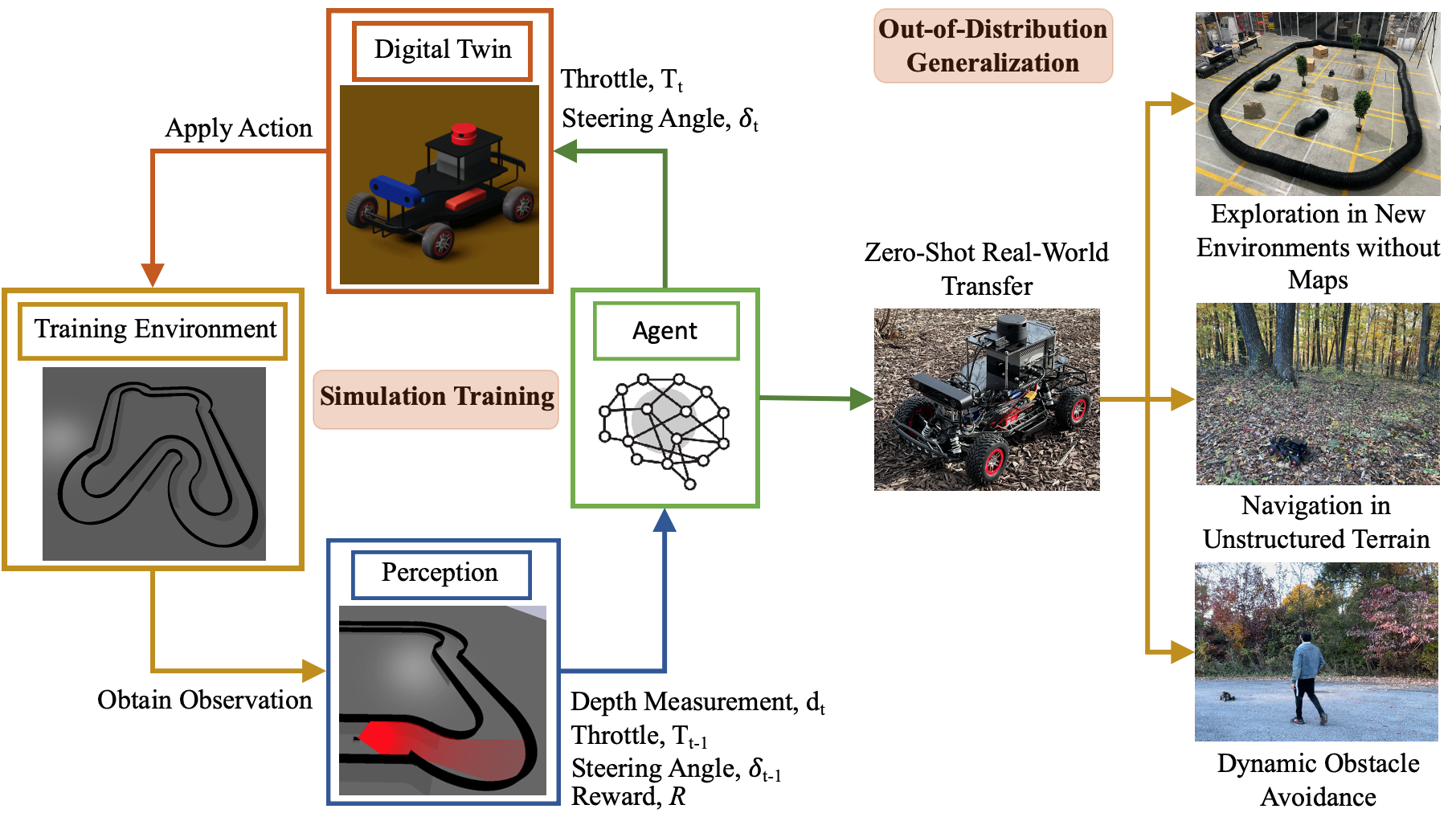}
    \caption{Methodology for training and deployment. The DRL model was trained in simulation with a racetrack environment transition probability and transferred zero-shot to the real world for OOD spatial reasoning with generalization to exploration in new environments without maps that include unstructured terrain and dynamic obstacle avoidance. A video of the results can be accessed at \href{https://www.youtube.com/watch?v=jUtPaQV3Bd8}{https://www.youtube.com/watch?v=jUtPaQV3Bd8}.} \label{method}
\end{figure*}

We used the open-source XTENTH-CAR wheeled AMR (\cite{sivashangaran2023xtenth}) and AutoVRL simulator (\cite{sivashangaran2023autovrl}), developed in-house in previous work, for sim-to-real training and evaluation. We transferred the simulation trained policy zero-shot to the real world for OOD generalization to navigation in previously unseen environments using a combination of sensor processing to match simulation observations and a reward component to offset an observed difference between simulation and real world physics. Figure \ref{method} illustrates the training and deployment schematic.
 
\subsection{Observation Space} \label{se:method_observation_space}

In order to facilitate quick policy learning and efficient real-time execution, we utilized depth observations that can be obtained using LiDAR or camera RGB-D point cloud. These measurements provide a complete spatial understanding of the surrounding environment at a fraction of the data size of RGB images, akin to animals that use echolocation as a result of natural evolution, such as bats (\cite{jones2007bat}). Furthermore, the domain gap between simulation and real world depth measurements is substantially smaller than for images, facilitating more accurate end-to-end policy transfer, without separate high-level feature extraction.

To improve computational efficiency and facilitate feasible training using consumer computer hardware, we utilized a 1D observation space $\mathcal{O}$, defined as follows, where $r_{i}$ is the depth measurement at each angle increment in the observation Field of View (FoV). The observations comprise 170 depth measurements spread across a frontal $120^{\circ}$ FoV.

\begin{equation} \label{}
\mathcal{O} = [r_{i}]^{1\times170}
\end{equation}

\subsection{Action Space} \label{se:method_action_space}

We define the 2D vector action space \textit{a} for the Ackermann steered AMR used for evaluation as follows.

\begin{equation} \label{}
\textit{a} = (a_{T}, a_{\delta});\;\;\;\; a_{T}, a_{\delta} \in [-1, 1]
\end{equation}\

The DRL algorithm outputs normalized continuous actions $a_{T}$ and $a_{\delta}$ in the range [-1 1]. We convert these to high-level control inputs, throttle $T$ in the range [0 1], and steering angle $\delta$ in the range $[\delta_{min}  \delta_{max}]$ as follows.

\begin{equation} \label{}
T = min(max(a_{T}, 0), 1)
\end{equation}

\begin{equation} \label{}
\delta = max(min(a_{\delta}, \delta_{max}), \delta_{min})
\end{equation}

In simulation, we applied the control inputs to the digital twin's joints using the physics engine's motor control function as detailed in (\cite{sivashangaran2023autovrl}). During real world inference, the throttle commands were linearly mapped to linear velocity. The steering inputs were not further processed. Low-level propulsion and servo motor RPMs were computed from these high-level actions and implemented using an open-source electronic speed controller. 

For AMRs that are differential wheeled or legged such as quadrupeds and humanoids, equipped with low-level controllers to control wheel speed or joint positions that track linear and angular velocities, the latter high-level commands may be used to replace throttle and steering angle actions.

\subsection{Reward Formulation} \label{se:method_reward}

We evaluated shaped and sparse rewards, and found the latter to perform best. Shaped rewards, identical to and variations of those in (\cite{sivashangaran2023deep}), penalized collisions and states near racetrack boundaries while rewarding throttle output, which caused excessive wall avoidance that slowed lap times, and overfitting that limited exploratory behavior. The optimal sparse reward in simulation was the squared throttle output $R=T^2$, equivalent to linear velocity for homogeneity with other AMR form factors, chosen for fast convergence, without an explicit collision penalty but episode termination upon boundary contact, enabling policies that maximized velocity, avoided overfitting on collision avoidance, and parameterized spatial reasoning for OOD generalization.

However, this policy failed in the real world, producing zig-zag trajectories from exploiting physics engine imperfections via rapidly oscillating steering angles during accelerated training, that was infeasible with real world sensor and actuator sampling rates. To address this, a revised reward was formulated as follows, penalizing negative products of previous and current unprocessed steering angles at the upper bound, which maintained simulation training time and performance but enabled zero-shot real world transfer without domain randomization or modeled sensor and actuator noise, with weights prioritizing throttle maximization.

\begin{equation} \label{eqn_reward_f}
    R = 
\begin{dcases}
    5T^{2} &\\ 
    -2.0              & \text{if } a_{\delta,prev}a_{\delta,curr} = -1\\
\end{dcases}
\end{equation}

\subsection{Training Process} \label{se:method_train}

A training episode was concluded when the agent collided with the racetrack boundary, following which it was reset to the same starting position. We trained the model with an Intel Core i9 13900KF CPU and NVIDIA GeForce RTX 4090 GPU, accelerated at 30 times real-time speed, for 20,000,000 steps which corresponded to 15,747 training episodes, in 48 wall-clock hours. The training hyperparameters are summarized in Table \ref{parameters}.

\begin{table}[ht]
    \renewcommand{\arraystretch}{1.2}
    \centering
    \caption{Training Hyperparameters}
    \label{parameters}
        \begin{tabular}[t]{|c|c|}
            \hline
            \textbf{Hyperparameter} & \textbf{Value} \\
            \hline
            Number of Epochs & 10 \\
            \hline
            Batch Size & 64 \\
            \hline
            Rollout Buffer Size & 2048 \\
            \hline
            Discount Factor ($\gamma$)  & 0.99 \\
            \hline
            Learning Rate & 0.0003 \\
            \hline
        \end{tabular}
\end{table}

\section{RESULTS AND DISCUSSION} \label{se:results}

This section presents and analyzes the effectiveness and limitations of the proposed method in different simulated and real world scenarios. In simulation, we evaluated time-optimal motion planning in the racetrack used for training, and tested OOD exploration capabilities in a densely cluttered 50$m$ x 50$m$ environment, with comparison of sample efficiency and environment converge to DRL approaches trained directly in the test environment. Real world experiments include exploration in a controlled object-filled laboratory environment, navigation in unstructured forest terrain and dynamic pedestrian avoidance.   

\subsection{Training Environment} \label{se:results_perf_training_env}

\begin{figure}[!h]
    \centering
    \includegraphics[width = 0.7\columnwidth]{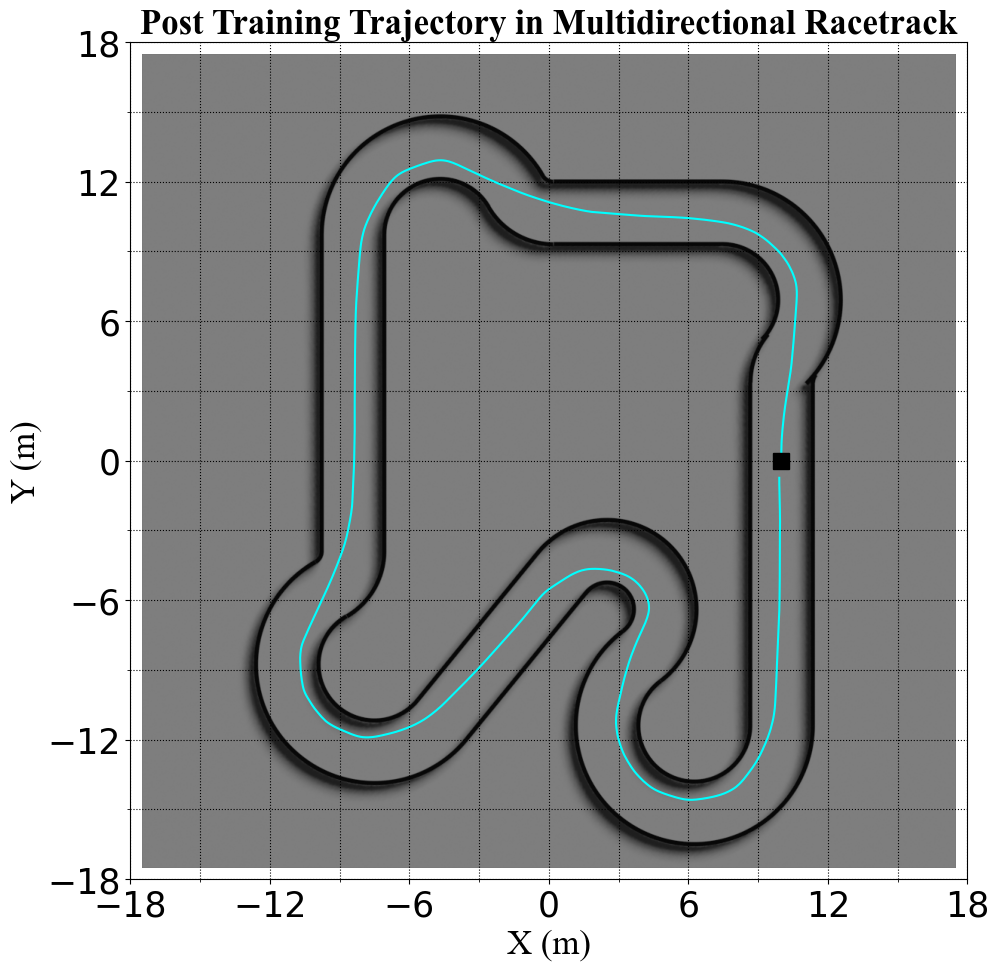}
    \caption{Time-optimal trajectory in the multidirectional racetrack used for training.} \label{racetrack_sim_trajectory}
\end{figure}

The post-training trajectory in the racetrack used for training is depicted in Figure \ref{racetrack_sim_trajectory}, where the trained model explored multiple turns of varying directions and radii without collision, time-optimally with a trajectory similar to professional racing drivers. The AMR traversed the apex of each corner to minimize distance traveled and thus lap time, requiring prediction and planning to execute sequential control signals that maximize velocity at physical limits in nonlinear domains.

The computationally efficient ANN with 2 hidden layers required significantly fewer operations than conventional racing algorithms that use optimization and sampling methods such as Model Predictive Control (MPC) and Rapid Random Trees (RRT) (\cite{hartmann2022competitive}). Moreover, the end-to-end DRL model does not require additional perception processing to extract explicit track boundary information, further increasing efficiency.

\subsection{Navigation in Simulated Obstacle-Filled Environment} \label{se:results_sim_out50}

\begin{figure}[!h]
    \centering
    \includegraphics[width = \columnwidth]{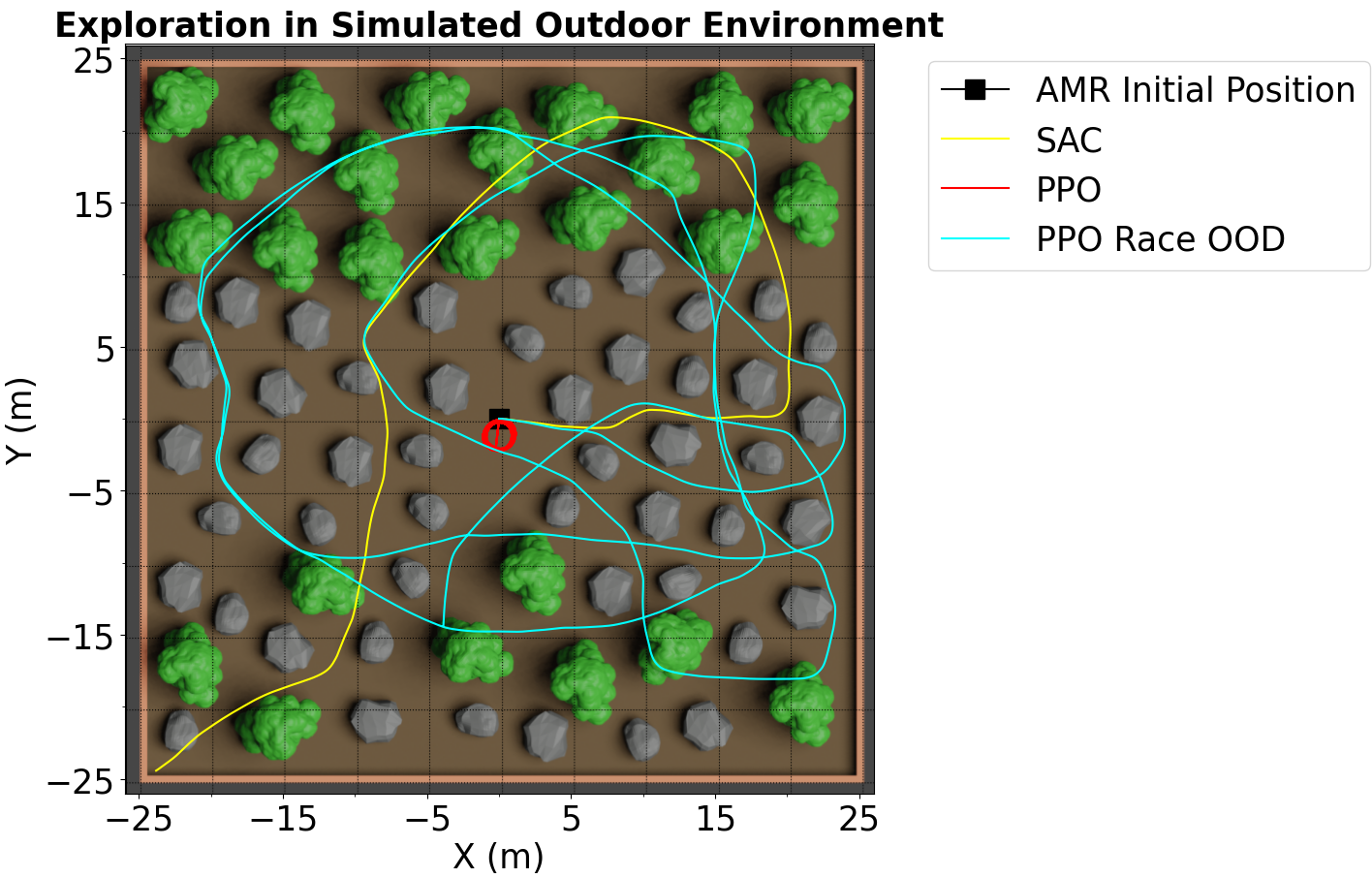}
    \caption{Exploration in a simulated outdoor environment with tree and boulder obstacles.} \label{outdoor_sim_trajectory}
\end{figure}

We evaluated the OOD obstacle avoidance and intrinsic exploratory performance of the ANN, trained to race, in a 50$m$ x 50$m$ simulated outdoor environment with tree and boulder obstacles, as illustrated in Figure \ref{outdoor_sim_trajectory}, without additional training or fine-tuning for navigation in obstacle-filled environments. We compared the exploration coverage of the proposed method to DRL models trained directly in this environment using PPO and Soft Actor-Critic (SAC) (\cite{haarnoja2018soft}), with the shaped reward in (\cite{sivashangaran2023deep}) specifically designed for exploration without maps. 

The OOD method, trained to race in 20,000,000 samples, explored this environment more effectively than policies trained in the test environment over 60,000,000 samples. The results are summarized in Table \ref{sim_env_exploration}. The exploration coverage was determined by the number of discretized 10 $m^2$ grid cells occupied by a trajectory. We chose 10 $m^2$ since onboard sensors such as LiDARs and multiple depth cameras can obtain viable 360$^{\circ}$ FoV data within this area.

\begin{table}[ht]
    \renewcommand{\arraystretch}{1.2}
    \centering
    \caption{Comparison of RL Training Methods}
    \label{sim_env_exploration}
        \begin{adjustbox}{max width=\columnwidth}
        \begin{tabular}[t]{|c|c|c|}
            \hline
            \textbf{Method} & \textbf{Training Samples} & \textbf{Exploration Coverage \%}\\
            \hline
            SAC  & 6$e^7$ & 52\\
            \hline
            PPO & 6$e^7$ & 8\\
            \hline
            \textbf{OOD PPO} & \textbf{2$\textbf{e}^7$} & \textbf{88}\\
            \hline
        \end{tabular}
        \end{adjustbox}
        
\end{table}

The OOD PPO policy explored 88\% of the environment compared to 52\% and 8\% for SAC and PPO trained directly in the test environment, where SAC outperformed PPO as the latter learned inefficient circular motion in the same vicinity despite a reward penalty for excessive steering input. In contrast, PPO learned a better policy with the transition probability in the racetrack environment due to the more pronounced alterations in the observation space from the track boundaries in comparison to sparsely detected objects. SAC took longer to converge in this environment, and did not learn the time-optimal trajectory due to more emphasis on exploratory behavior as a consequence of the algorithm's compromise between maximizing cumulative long-term reward and entropy, which led to conservative racetrack boundary avoidance that did not parameterize spatial reasoning as well as PPO.

\subsection{Zero-Shot Real World Transfer} \label{se:results_unstructured}

\begin{figure}[!h]
    \centering
    \includegraphics[width = 0.5\columnwidth]{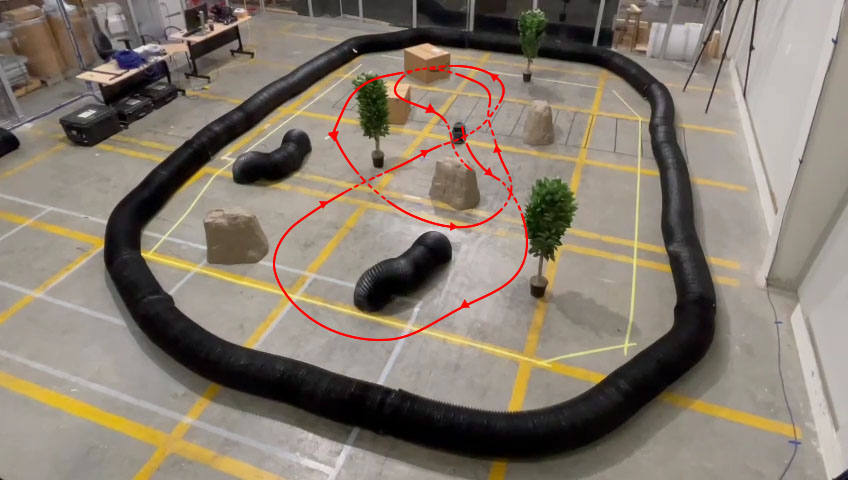}
    \caption{Exploration trajectory in a static object filled laboratory environment.} \label{laboratory_obstacle_traj}
\end{figure}

We transferred the simulation trained policy zero-shot to the physical AMR and evaluated its collision-free exploration performance in laboratory and outdoor environments. In the laboratory, the policy navigated a cluttered smooth-surfaced environment with diverse objects like boulders, tubes, and potted plants, as shown in Figure \ref{laboratory_obstacle_traj}, demonstrating high precision obstacle avoidance and sufficient exploration for onboard sensor data collection in most regions, matching simulation performance. 

\begin{figure}[htbp]
\centering
\begin{minipage}{0.252\textwidth}
\centering
\includegraphics[width=0.8\textwidth]{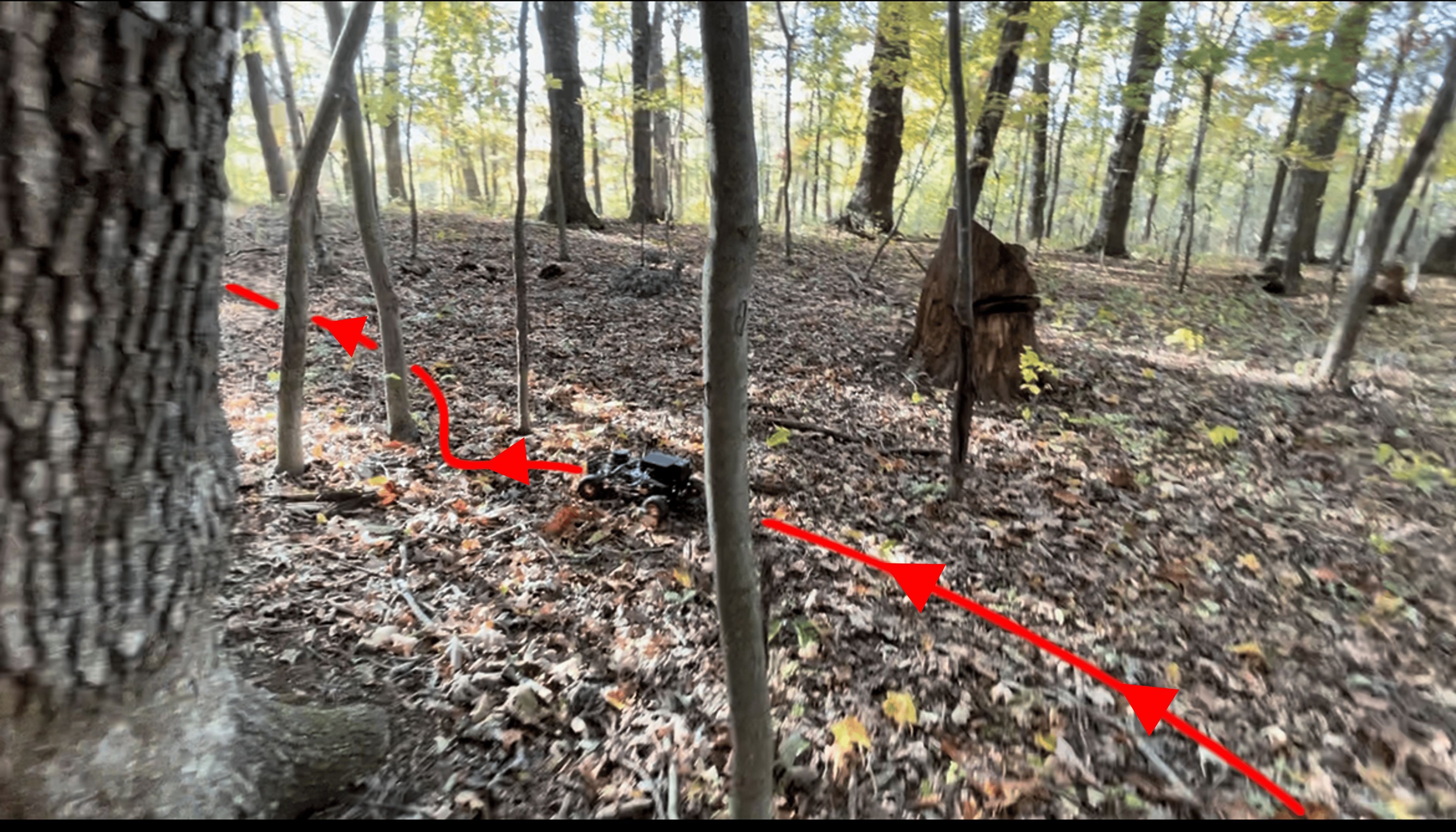}
\subcaption{} \label{forestnav1}
\end{minipage}%
\begin{minipage}{0.252\textwidth}
\centering
\includegraphics[width=0.8\textwidth]{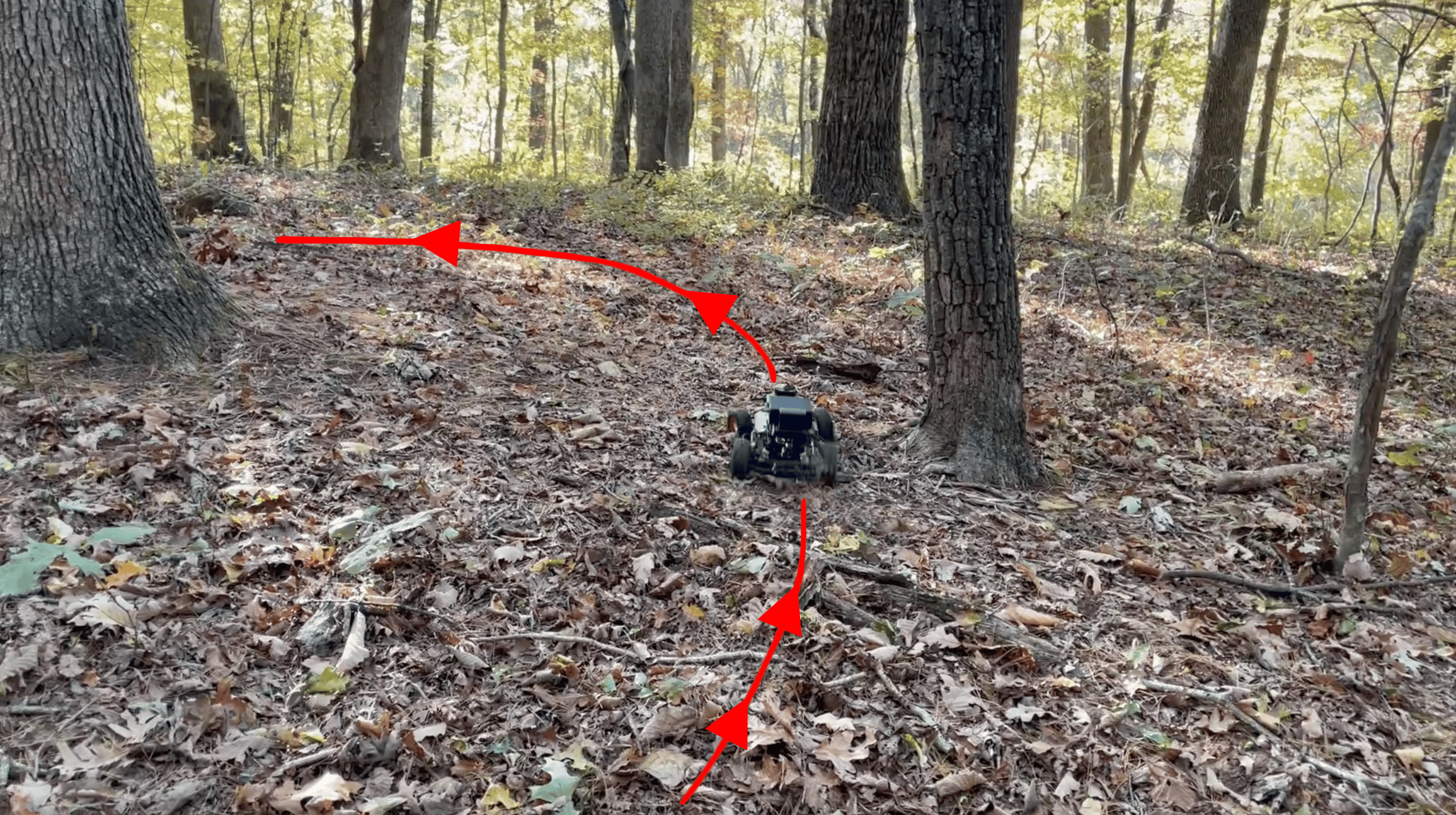}
\subcaption{} \label{forestnav3}
\end{minipage}

\caption{Exploration in unstructured forest terrain. (a) Navigation in narrow region between sparsely detected slender trunks. (b) Obstacle seeking behavior to tree trunks that is beneficial for exploration of aerially occluded shaded regions.} \label{forestnav}
\end{figure}

In the outdoor unstructured forest terrain with vegetation, soil, and leaf litter, as illustrated in Figure \ref{forestnav}, the AMR showed resilience to increased sensor and actuator noise from terrain elevation changes, accurate obstacle avoidance including in narrow regions between sparsely detected slender trunks, and qualitatively efficient trajectories without repeated coverage. Moreover, racing parameterized spatial reasoning enabled close proximity navigation beneficial for exploring aerially occluded shaded regions, enhancing ground AMR search capabilities. 

The absence of computationally expensive SLAM algorithms and mandatory high-level planners for goal positions enables greater adaptability, safety, and efficiency across environments and applications, particularly for initial exploration and data collection to build detailed maps for goal-oriented downstream tasks, while supporting high-speed real-time execution on AMRs with varying embedded compute payloads. In Table \ref{compute}, we summarize the CPU and memory usages on a state-of-the-art NVIDIA Jetson Orin AGX embedded computer, for the DRL model and conventional motion planning algorithms (\cite{khairnar2023comparison}) with computationally efficient 2D SLAM (\cite{KohlbrecherMeyerStrykKlingaufFlexibleSlamSystem2011}) and user provided goal positions.

\begin{table}[ht]
    \renewcommand{\arraystretch}{1.2}
    \centering
    \caption{CPU and Memory Usage}
    \label{compute}
        \begin{tabular}[t]{|c|c|c|}
            \hline
            \textbf{Method} & \textbf{CPU \%} & \textbf{Memory \%}\\
            \hline
            SLAM + pRRT  & 22.4 & 14.3\\
            \hline
            SLAM + MPC & 21.3 & 14.2\\
            \hline
            SLAM + A* & 21.6 & 14.3\\
            \hline
            \textbf{End-to-End DRL} & \textbf{7.6} & \textbf{12.6}\\
            \hline
        \end{tabular}
\end{table}

Our end-to-end DRL approach utilizes a fraction of the CPU resources, and lower memory. The difference in computation efficiency will be greater with 3D LiDAR and vision based SLAM algorithms.

\subsection{Dynamic Obstacle Avoidance} \label{se:results_dynamic}

\begin{figure}[htbp]
\centering
\begin{minipage}{0.16\textwidth}
  \centering
\includegraphics[width=0.9\textwidth]{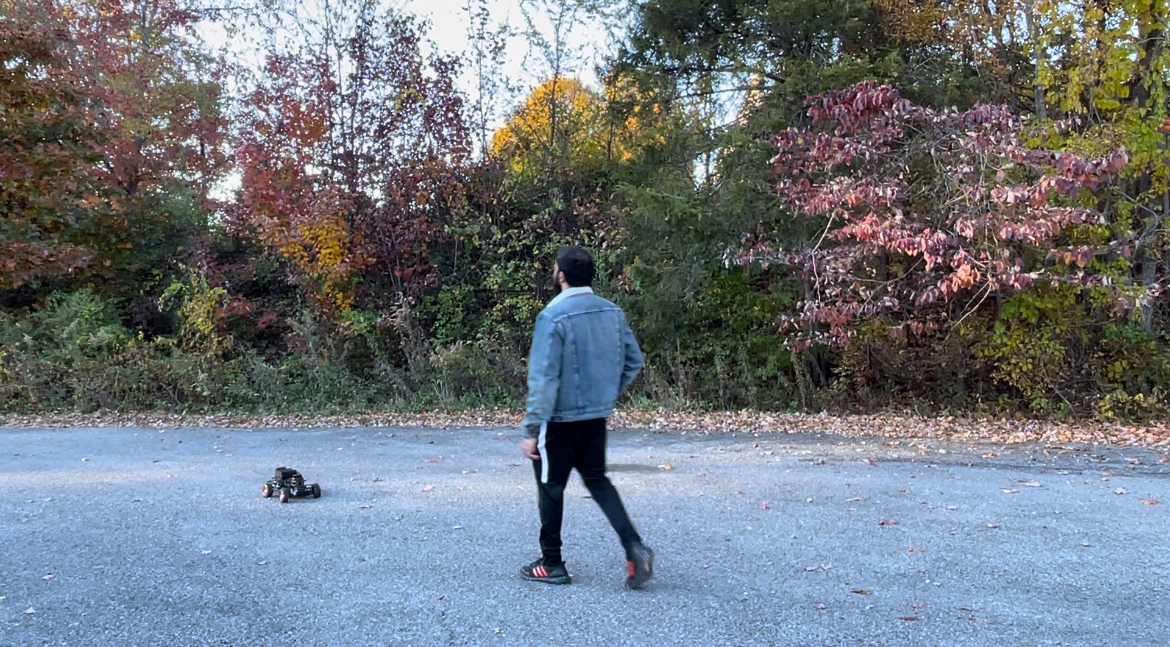}
\subcaption{} \label{dynamic_2_1}
\end{minipage}%
\begin{minipage}{0.16\textwidth}
  \centering
\includegraphics[width=0.9\textwidth]{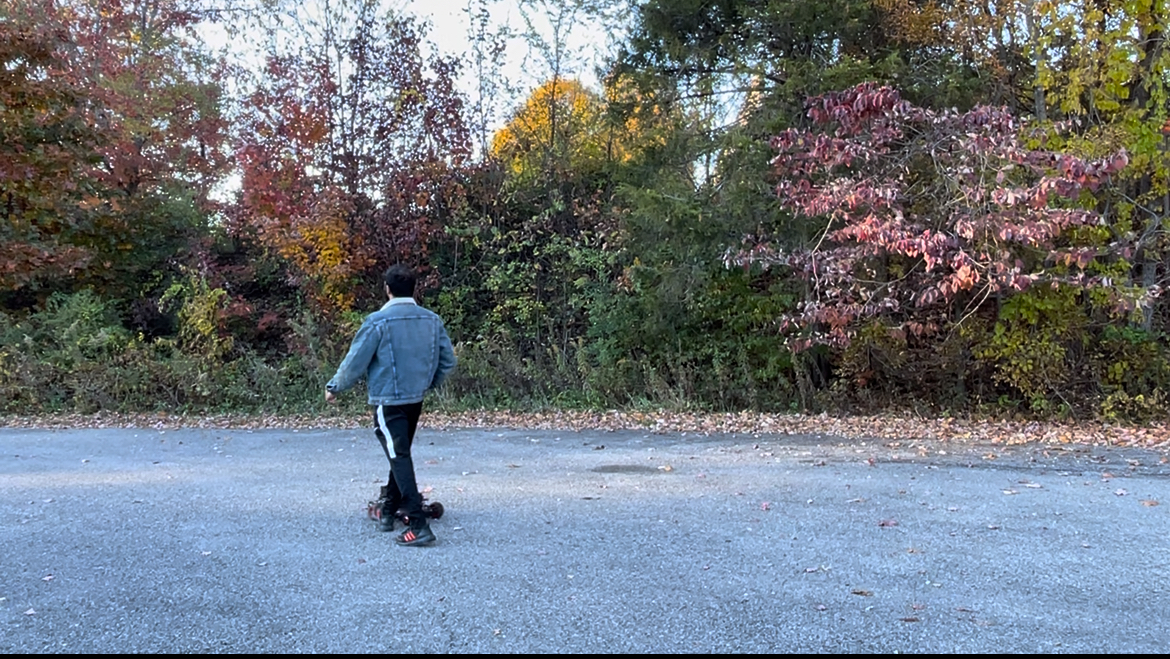}
\subcaption{} \label{dynamic_2_2}
\end{minipage}
\begin{minipage}{0.16\textwidth}
  \centering
\includegraphics[width=0.9\textwidth]{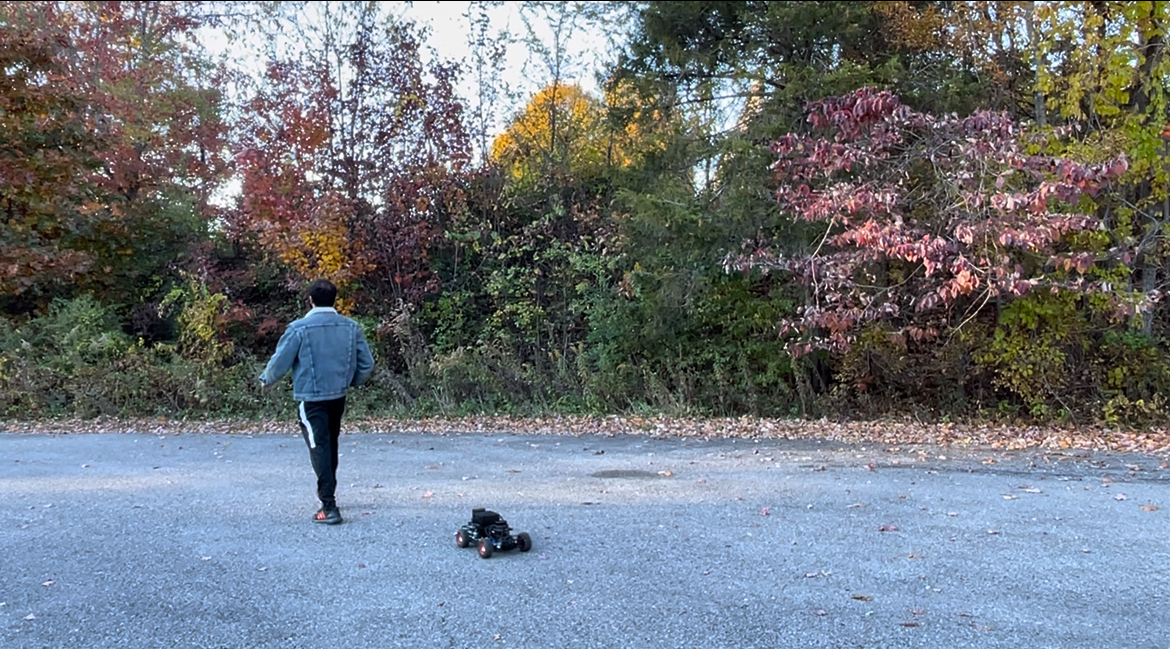}
\subcaption{} \label{dynamic_2_3}
\end{minipage}%

\caption{Dynamic pedestrian avoidance (a) Trajectories on collision course. (b) Evasive maneuver. (c) Dynamic obstacle avoided.} \label{dynamic_1}
\end{figure}

The prediction and planning required to execute control inputs in sequence, based on detected object profiles, to maximize velocity for racing at nonlinear physical limits, facilitated generalization to dynamic obstacle avoidance. We evaluated this capability to avoid pedestrian movement, at an operating speed of 1 $m/s$ in different collision directions. An example scenario is shown in Figure \ref{dynamic_1}.

\section{CONCLUSIONS}

AMR utility in dynamic, real world environments without detailed maps is an unsolved problem that has potential to improve humanity's capabilities across a variety of tasks. This paper presented a sample efficient DRL method to parameterize spatial reasoning in a compact, computationally efficient ANN with a simulated racetrack environment transition probability, for zero-shot OOD exploration and navigation applications. We used a reward component to mitigate differences between simulation and real world physics, and utilized depth measurement observations for fast inference, accurate simulation to real world policy transfer, and operation over a wide range of environments that include those that are GPS denied and dimly lit such as caves, bunkers and forests at night. The approach is robust to sensor and actuator noise, enabling operation in unstructured forest terrain, and capable of dynamic obstacle avoidance. Moreover, the method utilizes a fraction of the computation resources required for conventional modular navigation approaches, enabling execution in a range of AMR forms with varying onboard computer payloads.

\bibliography{ifacconf}

\end{document}